\documentclass{article}

\usepackage{arxiv}

\usepackage[utf8]{inputenc}
\usepackage[T1]{fontenc}
\usepackage{hyperref}
\usepackage{url}
\usepackage{booktabs}
\usepackage{amsfonts}
\usepackage{nicefrac}
\usepackage{microtype}
\usepackage{graphicx}
\usepackage{float}
\usepackage[numbers]{natbib}
\usepackage{amsmath,amssymb}
\usepackage{bm}
\usepackage{multirow}
\usepackage{subcaption}

\graphicspath{{Figures/}}

\newcounter{algorithm}
\newcounter{algline}
\newenvironment{algorithmblock}[1]{%
  \refstepcounter{algorithm}%
  \setcounter{algline}{0}%
  \par\smallskip\noindent\begin{minipage}{\linewidth}
  \hrule\kern2pt
  \noindent\textbf{Algorithm \thealgorithm: #1}\par
  \kern2pt\hrule\kern2pt
  \footnotesize
  \selectfont
  \setlength{\parskip}{0pt}%
}{%
  \kern2pt\hrule
  \end{minipage}\par\smallskip
}
\newcommand{\algtext}[1]{%
  \par\noindent #1\par
}
\newcommand{\algline}[2][0pt]{%
  \stepcounter{algline}%
  \par\noindent
  \makebox[1.8em][r]{\thealgline}\hspace{0.5em}\hspace*{#1}%
  \begin{minipage}[t]{\dimexpr\linewidth-2.3em-#1\relax}#2\end{minipage}
}

\title{Fermat Active Laplace Learning for Semi-Supervised Hyperspectral Image Classification}

\author{
  Vutichart Buranasiri\thanks{The authors gratefully acknowledge support from NSF DMS-2318894.} \\
  Department of Mathematics\\
  Tufts University\\
  Medford, MA 02155, USA \\
  \And
  James M.\ Murphy\footnotemark[1] \\
  Department of Mathematics\\
  Tufts University\\
  Medford, MA 02155, USA
}

\begin{document}
\maketitle

\begin{abstract}
Two active learning algorithms for hyperspectral image (HSI) classification are proposed that combine density-aware Fermat distances with Poisson-reweighted harmonic label propagation. Our methods actively query points using an uncertainty-based acquisition function, extending \textit{Poisson ReWeighted Laplace Learning} (PWLL). Our first algorithm, \textit{Fermat Active Laplace Learning} (FALL), builds an affinity matrix using Fermat distances between all data points. Then, PWLL is run with a diagonal perturbation using the minimum-norm acquisition function. In contrast, \textit{Approximate FALL} (A-FALL) computes Fermat distances between each data point and landmark pixels selected via farthest-point sampling and constructs the affinity matrix using landmark multidimensional scaling. After several query rounds, A-FALL selects the Fermat exponent $p$ using a leave-one-out cross-validation variant. FALL and A-FALL leverage Fermat distances and subsequent harmonic label propagation to provide a density-aware estimation of the data manifold, improving labeling accuracy. Experiments on Salinas A and Pavia show the effectiveness of FALL and the scalability of A-FALL to large HSI scenes.
\end{abstract}


\section{Introduction}

Machine learning has led to major advances in remote sensing \cite{wang2022self,li2024deep,ahmad2025comprehensive}; however, many modern approaches require a large number of labeled training samples. While recent supervised deep learning architectures can achieve superior labeling accuracy on benchmark hyperspectral images (HSI) \cite{gic_hsi_scenes}, they are often impractical when access to training data is limited. This has motivated semi-supervised methods that require only a small subset of labeled data for HSI classification \cite{jia2021survey}. In particular, graph-based semi-supervised methods such as Laplace learning \cite{zhu2003semi} and Poisson learning \cite{calder2020poisson} have shown strong performance on benchmark high-dimensional datasets.

Recently, the active learning \cite{cohn1995active,balcan2007margin,castro2008minimax,dasgupta2008general,balcan2009agnostic,dasgupta2011two} regime has shown promise for HSI segmentation \cite{rajan2008active,li2010semisupervised,demir2011batch,di2011multiview,tuia2011survey,patra2017spectral,haut2018active,murphy2018iterative,murphy2018unsupervised,deng2019active,maggioni2019active,murphy2020spatially,polk2022active,tripathi2024learning}. Starting from an unlabeled dataset, active learning iteratively queries points for labels to select the most informative pixels. There is a vast literature on query selection \cite{settles2009active}. Broadly, such methods can be categorized into those that prioritize exploring unknown regions of the dataset and those that exploit the downstream classifier's decision boundaries \cite{dasgupta2011two}.

The \textit{Poisson ReWeighted Laplace Learning} (PWLL) algorithm combines an uncertainty sampling querying method with a variant of the properly weighted Laplace Learning classifier, showing strong results on certain datasets \cite{miller2023poisson}. A key algorithmic choice in graph-based methods is the construction of edge weights. While the RBF kernel is commonly used, our approach considers a family of weighted shortest-path metrics known as Fermat distances \cite{howard2001geodesics,alamgir2012shortest,hwang2016shortest,little2020path,groisman2022nonhomogeneous,little2022balancing,garciatrillos2024fermat}, which use powers of Euclidean distances as weights. Fermat and other shortest-path distances have been used in unsupervised and semi-supervised machine learning problems \cite{sajama2005estimating,bijral2011semi,mckenzie2019power,fernandez2023intrinsic,garciatrillos2024fermat,tan2026highdimensional}.

Motivated by these developments, we propose \textit{Fermat Active Laplace Learning} (FALL) and \textit{Approximate FALL} (A-FALL). FALL extends PWLL by using an affinity matrix built from Fermat distances over all data points. In contrast, A-FALL computes Fermat distances over $m$ landmark pixels and uses landmark multi-dimensional scaling (LMDS) \cite{desilva2002global} to compute approximate weights for downstream PWLL classification. Additionally, a variant of leave-one-out cross-validation (LOO-CV) \cite{zhang2006hyperparameter,wang2018approximate} is used in A-FALL to score a fixed set of candidate $p$-values. Experiments on Salinas A and Pavia show the effectiveness of FALL and the computational efficiency of A-FALL in the low-label rate regime.

The rest of this paper is organized as follows. In Section \ref{sec:Background}, we review relevant background material. We outline Fermat exponent learning algorithms in Section \ref{sec:plearning}, proposed algorithms in Section \ref{sec:algorithms}, and experimental results in Section \ref{sec:results}. Code is available at: \url{https://github.com/vburan01/FALL}.

\section{Background}\label{sec:Background}

\subsection{Fermat Distances}\label{subsec:Fermat}

Let $X=\{x_i\}_{i=1}^{N}\subset\mathbb{R}^{D}$ denote a HSI with $D$ spectral bands, with pixels $\{x_i\}_{i=1}^{N}$ in $\mathbb{R}^{D}$. Given pixels $x_i$ and $x_j$, the discrete $p$-weighted Fermat distance between $x_i$ and $x_j$ is

\[
\ell_p(x_i,x_j)
=
\min_{\pi=\{x_k\}_{k=1}^{T}}
\left(
\sum_{k=1}^{T-1}\|x_k-x_{k+1}\|^p
\right)^{1/p}
\]

where $\pi$ is a path in $\{x_i\}_{i=1}^{N}$ such that $x_1=x_i$ and $x_T=x_j$, and $\|\cdot\|$ is the Euclidean 2-norm. Intuitively, $\ell_p(x_i,x_j)$ is small when $x_i$ and $x_j$ are close in the geodesic sense, and the shortest path passes through high-density regions. As $p\geq1$ increases, paths through high-density regions are favored more than paths through points close in geodesic distance.

Let $G_X^p$ denote the complete graph with edge weights $\|x_i-x_j\|^p$, and let $G_X^{p,k}\subset G_X^p$ be the undirected, union-symmetrized $k$NN subgraph that retains an edge $\{x_i,x_j\}$ if $x_i$ is one of the $k$ nearest neighbors of $x_j$, or vice versa. It is shown in \cite{little2022balancing} that if $k\approx\log(N)$, the sample Fermat distances computed over $G_X^p$ and $G_X^{p,k}$ agree with high probability under standard sampling assumptions. Thus, Fermat distances can be computed using Dijkstra's algorithm in $\mathcal{O}(N^2\log^2(N))$ time instead of $\mathcal{O}(N^3)$.

\subsection{Poisson Reweighted Laplacian Learning}\label{subsec:PWLL}

We summarize the PWLL-$\tau$ method introduced in \cite{miller2023poisson}. Let $X=\mathcal{U}\cup\mathcal{L}$, where $\mathcal{L}$ is the set of labeled data points and $\mathcal{U}$ are unlabeled. Let $C$ be the number of classes, and $e_i\in\mathbb{R}^{C}$ be the $i^{\mathrm{th}}$ canonical basis vector. The PWLL-$\tau$ algorithm solves the following problem for a function $u:X\rightarrow\mathbb{R}^{C}$:

\[
\min_u
\sum_{x_i,x_j\in X}
\frac{1}{2}\gamma(x_i)\gamma(x_j)W_{ij}
\|u(x_i)-u(x_j)\|^2
+
\tau\sum_{x_i\in\mathcal{U}}\|u(x_i)\|^2
\]

subject to $u(x)=e_{y(x)}$ for $x\in\mathcal{L}$, where $y(x)$ denotes the class label of $x$, $W_{ij}$ is the $(i,j)^{\mathrm{th}}$ entry of the chosen symmetric, nonnegative weight matrix $W$, and $\tau\geq0$ controls the weighting of the diagonal perturbation. Larger $\tau$ encourages $u$ to take smaller values for data points away from the labeled set $\mathcal{L}$. In practice, we apply an exponential decay schedule for $\tau$ over active learning rounds. The reweighting function $\gamma:X\rightarrow\mathbb{R}_{+}$ is found by solving the graph Poisson equation,

\[
\sum_{x_j\in X}W_{ij}\big(\gamma(x_i)-\gamma(x_j)\big)
=
\sum_{x_k\in\mathcal{L}}\left(\delta_{ik}-\frac{1}{N}\right)
\]

for all $x_i\in X$, where $\delta_{ik}$ is the Kronecker delta. We resolve the additive ambiguity of solutions by grounding one node and uniformly shifting the resulting potential so its minimum value is strictly positive. Thus, the reweighting $\gamma$ is learned from the data and is motivated by the continuum version of the graph Poisson equation, whose fundamental solution produces the appropriate scaling of $\gamma$ in regions near the labeled set $\mathcal{L}$ \cite{miller2023poisson}. The inferred classification $\hat{y}(x_i)$ for a pixel $x_i$ is given by $\hat{y}(x_i)=\arg\max_{k\in\{1,\ldots,C\}}u_k(x_i)$, where $u_k(x_i)$ is the $k^{\mathrm{th}}$ entry of $u(x_i)\in\mathbb{R}^{C}$.

After each PWLL-$\tau$ solve, the minimum-norm acquisition function $A:X\rightarrow\mathbb{R}_{\geq0}$, given by $A(x_i)=\|u(x_i)\|$ for $x_i\in X$, is used to iteratively query the next labeled point. After $K$ querying rounds, the $(K+1)^{\mathrm{th}}$ queried point is given by $\arg\min_{x_j\in\mathcal{U}}A(x_j)$. Intuitively, $A$ reflects the uncertainty of the classification produced by the PWLL-$\tau$ algorithm. With decaying $\tau$-regularization, points selected by $A$ are encouraged to spread over unexplored regions until sufficient labeling has occurred to cover the dataset in the low-label rate regime. Afterward, as $\tau\rightarrow0$, $A$ will take on smaller values in regions of high uncertainty. Thus, the acquisition function $A$ effectively balances exploration and exploitation of data points for querying.

\subsection{Landmark Multi-dimensional Scaling}\label{subsec:LMDS}

Given data points $\{x_i\}_{i=1}^{N}$, let $\{z_i\}_{i=1}^{m}$ denote a set of landmark points, where $m<N$. We use farthest point sampling (FPS) \cite{eldar1997farthest} to select landmark points, which iteratively chooses the $(k+1)^{\mathrm{th}}$ pixel furthest in spectral $\ell_2$ norm from the previous $k$ points. Let $D^{\mathrm{dist}}\in\mathbb{R}^{N\times m}$ be the distance matrix defined by $D_{ij}^{\mathrm{dist}}=\ell_p(x_i,z_j)$, and let $\Delta_m$ be the squared landmark-to-landmark sub-matrix defined by $(\Delta_m)_{ij}=\ell_p(z_i,z_j)^2$. LMDS is run using distances from $D^{\mathrm{dist}}$ as follows. We form the mean-centered inner product matrix 

\[B=-\frac12H\Delta_mH,\]

where $H$ is the centering matrix given by $H_{ij}=\delta_{ij}-1/m$. Let $\{\lambda_i\}_{i=1}^{n_p}$ denote the $n_p$ strictly positive eigenvalues of $B$ ordered such that $\lambda_1\geq\lambda_2\geq\cdots\geq\lambda_{n_p}$, and let $v_i$ denote corresponding orthonormal eigenvectors associated with $\lambda_i$ for each $i=1,\ldots,n_p$. For a chosen embedding dimension $r\leq n_p$, the landmark embedding matrix $L$ is given by $L=[\sqrt{\lambda_1}v_1,\sqrt{\lambda_2}v_2,\ldots,\sqrt{\lambda_r}v_r]\in\mathbb{R}^{m\times r}$. The $i^{\mathrm{th}}$ row of $L$ gives the $r$-dimensional LMDS embedding coordinates of $z_i$, denoted $\{z_i^{MDS}\}_{i=1}^{m}$. To obtain the remaining embedding coordinates for non-landmark points, let $\Delta_{x_i}\in\mathbb{R}^{m}$ be defined entrywise by $(\Delta_{x_i})_j=\ell_p(x_i,z_j)^2$ for each pixel $x_i$. Define $\bar{\Delta}_m\in\mathbb{R}^{m}$ componentwise by 

\[(\bar{\Delta}_m)_j=\frac{1}{m} \sum_{b=1}^{m}\ell_p(z_j,z_b)^2.\]

The embedding coordinates of $x_i$ are given by $x_i^{MDS}=L^{\dagger}(\bar{\Delta}_m-\Delta_{x_i})/2\in\mathbb{R}^{r}$, where $L^{\dagger}$ is the pseudoinverse of $L$, given by $L^{\dagger}=[v_1/\sqrt{\lambda_1},v_2/\sqrt{\lambda_2},\ldots,v_r/\sqrt{\lambda_r}]^{\top}\in\mathbb{R}^{r\times m}$.

\subsection{Kron Reduction}\label{subsec:Kron}

We now develop the application of Kron reduction to the PWLL harmonic label propagation problem, following the framework of \cite{dorfler2013kron}. We first note that the minimization in Section \ref{subsec:PWLL} is equivalent to minimizing the $\tau$-regularized Dirichlet energy of the Poisson Reweighted Laplacian $L_\gamma=D_\gamma-W_\gamma$, where $W_\gamma$ is the reweighted weight matrix given entrywise by $(W_\gamma)_{ij}=\gamma_i\gamma_jW_{ij}$, and $D_\gamma$ is the reweighted diagonal degree matrix given by $(D_\gamma)_{ii}=\sum_j(W_\gamma)_{ij}$. Let $U\in\mathbb{R}^{N\times C}$ be the matrix obtained via stacking $u(x_i)^\top$ row-wise. We can block partition $L_\gamma$ and $U$ into labeled and unlabeled components:

\[
L_\gamma=
\begin{bmatrix}
L_{\mathcal{LL}}&L_{\mathcal{LU}}\\
L_{\mathcal{UL}}&L_{\mathcal{UU}}
\end{bmatrix}
\quad\text{and}\quad
U=
\begin{bmatrix}
U_{\mathcal{L}}\\
U_{\mathcal{U}}
\end{bmatrix}.
\]

Note that since 

\[\sum_{x_i,x_j\in X}\frac12(W_\gamma)_{ij}\|u(x_i)-u(x_j)\|^2=\operatorname{tr}(U^\top L_\gamma U),\] 

it follows that the $\tau$-regularized harmonic problem is equivalent to

\[
\min_{U_{\mathcal{U}}}E(U_{\mathcal{L}},U_{\mathcal{U}})
=
\min_{U_{\mathcal{U}}}
\frac{1}{2}\operatorname{tr}(U^\top L_\gamma U)
+
\frac{\tau}{2}\operatorname{tr}(U_{\mathcal{U}}^\top U_{\mathcal{U}})
\]

subject to the constraint $U_{\mathcal{L}}=Y_{\mathcal{L}}$, where $Y_{\mathcal{L}}\in\mathbb{R}^{|\mathcal{L}|\times C}$ is the label matrix whose $i^{\mathrm{th}}$ row is given by $e_{y(x_i)}^\top$ for each labeled point $x_i\in\mathcal{L}$. Given that $L_{\mathcal{UU}}+\tau I$ is invertible, where $I$ is the $|\mathcal{U}|\times|\mathcal{U}|$ identity, taking the derivative with respect to $U_{\mathcal{U}}$ yields the optimal $U_{\mathcal{U}}$, given by 

\[U_{\mathcal{U}}^\star=-(L_{\mathcal{UU}}+\tau I)^{-1}L_{\mathcal{UL}}Y_{\mathcal{L}}.\] 

We note that $L_{\mathcal{UU}}+\tau I$ is invertible when $\tau>0$, and when $\tau=0$ and every connected component of the graph associated with $W_\gamma$ contains a labeled node. Substituting $U_{\mathcal{U}}^\star$ and simplifying yields the $\tau$-regularized reduced energy as a function of $Y_{\mathcal{L}}$:

\[
E(Y_{\mathcal{L}},U_{\mathcal{U}}^\star)
=
\frac{1}{2}\operatorname{tr}
\left(
Y_{\mathcal{L}}^\top
\left[
L_{\mathcal{LL}}
-
L_{\mathcal{LU}}(L_{\mathcal{UU}}+\tau I)^{-1}L_{\mathcal{UL}}
\right]
Y_{\mathcal{L}}
\right)
\]

where $S_\gamma=L_{\mathcal{LL}}-L_{\mathcal{LU}}(L_{\mathcal{UU}}+\tau I)^{-1}L_{\mathcal{UL}}$ is the Schur complement of the block matrix

\[
\begin{bmatrix}
L_{\mathcal{LL}}&L_{\mathcal{LU}}\\
L_{\mathcal{UL}}&L_{\mathcal{UU}}+\tau I
\end{bmatrix}
\]

with respect to $L_{\mathcal{UU}}+\tau I$. In particular, the quadratic form of $S_\gamma$ represents the reduced $\tau$-regularized Dirichlet energy over labeled nodes $\mathcal{L}$ after unlabeled nodes $\mathcal{U}$ have been eliminated through optimality conditions.

\section{Learning the Fermat Exponent}\label{sec:plearning}

The Fermat geometry is heavily dependent on the choice of $p$, which determines how density-aware the underlying data manifold is. We devise two methods, exact leave-one-out (ELOO) and approximate leave-one-out (ALOO), for learning $p$ during the active learning process, motivated by \cite{zhang2006hyperparameter,wang2018approximate}. ELOO exactly evaluates the ELOO cross-validation criterion for each candidate exponent in a fixed set, but it is expensive and impractical for larger HSI scenes and budgets. Thus, we propose ALOO, which uses principles of Kron reduction to learn $p$ efficiently.

\subsection{Exact LOO $p$-learning}\label{subsec:ELOO}

Suppose we have a set of candidate exponents $\mathcal{P}$ and a current labeled set $\mathcal{L}$. For each $a\in\mathcal{L}$ we consider leaving out $a$ and forming the reduced set $\mathcal{R}_a=\mathcal{L}\setminus\{a\}$. For each $p\in\mathcal{P}$, we form the Poisson reweighted weight matrix $W_{\gamma,\mathcal{R}_a}^{(p)}$ using labels in $\mathcal{R}_a$, defined by

\[\left(W_{\gamma,\mathcal{R}_a}^{(p)}\right)_{ij}=\gamma_i^{(p,\mathcal{R}_a)}\gamma_j^{(p,\mathcal{R}_a)}W_{ij}^{(p)}. \]

The weight matrix $W^{(p)}$ is formed using LMDS with landmark distances $\ell_p$, and $\gamma^{(p,\mathcal{R}_a)}$ is the solution of the graph Poisson equation with weights $W^{(p)}$ using labeled nodes $\mathcal{R}_a$. We then solve the PWLL-$\tau$ minimization subject to $u(x)=e_{y(x)}$ for all $x\in\mathcal{R}_a$ to obtain $u_{\mathcal{R}_a}^{(p)}:X\rightarrow\mathbb{R}^{C}$. The ELOO prediction for the held-out label $a$ is thus given by $u_{\mathcal{R}_a}^{(p)}(a)\in\mathbb{R}^{C}$. The ELOO algorithm solves the PWLL-$\tau$ minimization problem over $\mathcal{R}_a$ for each $a\in\mathcal{L}$ and for each $p\in\mathcal{P}$, leading to $|\mathcal{P}||\mathcal{L}|$ total solves. We then define the ELOO score function

\[
S_{\mathrm{ELOO}}(p)
=
\frac{1}{|\mathcal{L}|}
\sum_{a\in\mathcal{L}}
\left\|
\Pi\left(u_{\mathcal{R}_a}^{(p)}(a)\right)-e_{y(a)}
\right\|^2,
\]

where $\Pi:\mathbb{R}^{C}\rightarrow\mathbb{R}^{C}$ is a normalization function defined by

\[
\Pi(x)
=
\begin{cases}
    
\dfrac{[x]_{+}}{\sum_{i=1}^{C}[x_i]_{+}} & \text{If} \quad \sum_{i=1}^{C}[x_i]_{+}>\epsilon_{\mathrm{mach}},

\\ \\

\dfrac{1}{C}\mathbf{1} & \text{Otherwise.}

\end{cases}
\]

Where $\mathbf{1}$ is the all-ones vector, $\epsilon_{\mathrm{mach}}$ is machine precision, and $[x]_{+}=(\max\{x_1,0\},\ldots,\max\{x_C,0\})^\top$. Thus, $S_{\mathrm{ELOO}}$ calculates the average difference between normalized ELOO predictions and true labels. We introduce a confidence margin term,

\[
M_{\mathrm{ELOO}}(p)
=
\frac{1}{|\mathcal{L}|}
\sum_{a\in \mathcal{L}}\lambda_a(p),
\]

where

\[
\lambda_a(p)
=
\Pi\left(u_{\mathcal{R}_a}^{(p)}(a)\right)_{y(a)}
-
\max_{k\neq y(a)}
\Pi\left(u_{\mathcal{R}_a}^{(p)}(a)\right)_k.
\]

Hence, $M_{\mathrm{ELOO}}(p)$ finds the average margin between the true class score and the largest competing class score. The final margin-regularized ELOO score is given by 

\[\mathbf{S}_{\mathrm{ELOO}}(p)=\nu_1M_{\mathrm{ELOO}}(p)-S_{\mathrm{ELOO}}(p),\] 

where $\nu_1$ is a weight parameter.

\subsection{Approximate LOO $p$-learning}\label{subsec:ALOO}

In ALOO, we first form the Poisson reweighted weight matrix $W_{\gamma,\mathcal{L}}^{(p)}$ using all labels in $\mathcal{L}$ for each $p\in\mathcal{P}$, with associated Poisson reweighted Laplacian $L_{\gamma,\mathcal{L}}^{(p)}$. We then solve the PWLL-$\tau$ minimization subject to $u(x)=e_{y(x)}$ for all $x\in\mathcal{L}$ to obtain inferred classifications $u_{\star}^{(p)}:X\rightarrow\mathbb{R}^{C}$. Then, let $U_{\mathcal{U}}^\star\in\mathbb{R}^{|\mathcal{U}|\times C}$ denote the matrix with $i^{\mathrm{th}}$ row given by $u_{\star}^{(p)}(x_i)$, where $x_i$ is the $i^{\mathrm{th}}$ unlabeled data point in $\mathcal{U}$. Instead of exactly solving the PWLL-$\tau$ problem, we form the Schur complement

\[
S_p
=
\left(L_{\gamma,\mathcal{L}}^{(p)}\right)_{\mathcal{LL}}
-
\left(L_{\gamma,\mathcal{L}}^{(p)}\right)_{\mathcal{LU}}
\left[
\left(L_{\gamma,\mathcal{L}}^{(p)}\right)_{\mathcal{UU}}+\tau I
\right]^{-1}
\left(L_{\gamma,\mathcal{L}}^{(p)}\right)_{\mathcal{UL}}
\]

from Section \ref{subsec:Kron}. Then, for each $a\in\mathcal{L}$, we can define block-partitioned matrices from sets $\{a\}$ and $\mathcal{R}_a$ as

\[
F_{\mathcal{L}}
=
\begin{bmatrix}
F_a\\
Y_{\mathcal{R}_a}
\end{bmatrix}
\quad\text{and}\quad
S_p
=
\begin{bmatrix}
S_{p,aa}&S_{p,a\mathcal{R}_a}\\
S_{p,\mathcal{R}_aa}&S_{p,\mathcal{R}_a\mathcal{R}_a}
\end{bmatrix}.
\]

In particular, $F_a\in\mathbb{R}^{1\times C}$ is the unknown class prediction for $a$, and $Y_{\mathcal{R}_a}\in\mathbb{R}^{|\mathcal{R}_a|\times C}$ denote the matrix of fixed labels. When $a$ is held out, the $\tau$-regularized energy as a function of $Y_{\mathcal{L}}$ is given by

\[
E_a(Y_{\mathcal{L}},U_{\mathcal{U}}^\star)
=
\frac{1}{2}\operatorname{tr}
\left(
Y_{\mathcal{L}}^\top S_pY_{\mathcal{L}}
\right)
+
\frac{\tau}{2}\|F_a\|^2.
\]

It follows that minimizing $E_a(Y_{\mathcal{L}},U_{\mathcal{U}}^\star)$ is equivalent to minimizing

\[
E_a(F_a,Y_{\mathcal{R}_a})
=
\frac{1}{2}(S_{p,aa}+\tau)\|F_a\|_2^2
+
\operatorname{tr}(F_a^\top S_{p,a\mathcal{R}_a}Y_{\mathcal{R}_a})
+
K_{p,a},
\]

where $K_{p,a}=\frac{1}{2}\operatorname{tr}(Y_{\mathcal{R}_a}^\top S_{p,\mathcal{R}_a\mathcal{R}_a}Y_{\mathcal{R}_a})$ is a constant with respect to $F_a$. Hence, taking the first-order condition yields

\[(S_{p,aa}+\tau)F_a+S_{p,a\mathcal{R}_a}Y_{\mathcal{R}_a}=0.\] 

Solving for $F_a$ and finding ALOO predictions for $a$ thus amounts to computing

\[
F_{p,a}^{(\mathcal{R}_a)}
=
-
\frac{S_{p,a\mathcal{R}_a}Y_{\mathcal{R}_a}}
{S_{p,aa}+\tau}.
\]

ALOO thus computes predictions $F_{p,a}^{(\mathcal{R}_a)}$ for each $a\in\mathcal{L}$ and each $p\in\mathcal{P}$, without solving PWLL-$\tau$ for each set $\mathcal{R}_a$. Thus, the ALOO score function is
\[
S_{\mathrm{ALOO}}(p)
=
\frac{1}{|\mathcal{L}|}
\sum_{a\in\mathcal{L}}
\left\|
\Pi\left(F_{p,a}^{(\mathcal{R}_a)}\right)-e_{y(a)}
\right\|^2
\]
with the corresponding confidence margin term
\[
M_{\mathrm{ALOO}}(p)
=
\frac{1}{|\mathcal{L}|}
\sum_{a\in\mathcal{L}}\lambda'_a(p),
\]
where
\[
\lambda'_a(p)
=
\Pi\left(F_{p,a}^{(\mathcal{R}_a)}\right)_{y(a)}
-
\max_{k\neq y(a)}
\Pi\left(F_{p,a}^{(\mathcal{R}_a)}\right)_k.
\]
The final margin-regularized ALOO score value is hence given by 

\[\mathbf{S}_{\mathrm{ALOO}}(p)=\nu_2M_{\mathrm{ALOO}}(p)-S_{\mathrm{ALOO}}(p).\]

We use $\nu_1=\nu_2=0.02$ in experiments. The ALOO formulation assumes that $W_{\gamma,\mathcal{L}}^{(p)}\approx W_{\gamma,\mathcal{R}_a}^{(p)}$ for each $a\in\mathcal{L}$ and each $p\in\mathcal{P}$. Numerical experiments suggest that this property holds.

\section{The Proposed Algorithms}\label{sec:algorithms}

We propose two algorithms, FALL and A-FALL, which utilize Fermat distances in the underlying affinity matrix, leading to a density-aware estimation of the underlying data manifold. We further propose a method for learning the optimal exponent parameter $p$ in A-FALL using ALOO that is scalable to large HSI scenes (Section \ref{sec:plearning}). FALL and A-FALL show improved accuracy, with A-FALL being an efficient alternative.

A-FALL is outlined in Algorithm \ref{alg:AFALL}. The FPS algorithm \cite{eldar1997farthest} is first used to find $m$ landmark points, ensuring that a spectrally diverse set of points is chosen efficiently. We fix $m=300$ in experiments. Then, we compute Fermat distances over landmark points via shortest paths using Dijkstra's algorithm over a sparse $k$NN graph over $X$. The corresponding LMDS embeddings $\{x_i^{MDS}\}_{i=1}^{N}$ are computed according to Section \ref{subsec:LMDS}, where we set $r=32$. LMDS approximates Fermat distances efficiently, making Fermat graph construction and $p$-learning methods computationally feasible.

In A-FALL, $W^{(p)}$ uses LMDS distances $d_p(i,j)=\|x_i^{MDS}-x_j^{MDS}\|$ for the corresponding exponent $p$ and a self-tuned Gaussian kernel inspired by \cite{zelnikmanor2004self}. For each $x\in X$, we define the local bandwidth $\sigma_x=d_p(x,x_{k_\sigma})$ where $x_{k_\sigma}$ is the $k_\sigma^{\mathrm{th}}$ nearest neighbor of $x$ in the LMDS space. Given that $\mathcal{N}_{k_G}^{(p)}(i)$ is the set of $k_G$ nearest neighbors of $x_i$ excluding $x_i$ itself, $W^{(p)}$ is given entrywise by $W_{ij}^{(p)}=\max(\widehat{W}_{ij},\widehat{W}_{ji})$ and $W_{ii}^{(p)}=0$, where

\[\widehat{W}_{ij}
= \begin{cases}
\exp\left(-\frac{d_p(x_i,x_j)^2}{\eta^2\sigma_i\sigma_j}\right) & \text{if $j\in\mathcal{N}_{k_G}^{(p)}(i)$ and $j\neq i$,} \\ 
0 & \text{Otherwise.}
\end{cases}\]

 We set $k_\sigma=k_G=20$ and $\eta=8$ in experiments. The self-tuned kernel is used as it improves the numerical conditioning of the PWLL-$\tau$ solve. FALL is outlined in algorithm \ref{alg:FALL}. The weight matrix $W^{(p)}$ is defined entrywise in the same way as A-FALL, but with Fermat distances $d_p(i,j)=\ell_p(x_i,x_j)$. This leads to an exact representation of the Fermat geometry at the cost of higher computation time. In both algorithms, PWLL-$\tau$ is used as the downstream classifier as it mitigates the low-label regime degeneracy of Laplace learning through Poisson reweighting \cite{miller2023poisson}. Additionally, $\tau$-regularization increases numerical stability of the PWLL solve and ensures the minimum-norm acquisition function $A$ balances exploration and exploitation of data points. To ensure invertibility of $L_{\mathcal{UU}}+\tau I$, we require that the underlying graph associated with $W^{(p)}$ is connected. We use a decay schedule for $\tau$ as in \cite{miller2023poisson}, given by $\tau_t=\tau_0(\epsilon/\tau_0)^{t/(2C)}$ when $t<2C$ and $\tau_t=0$ when $t\geq2C$, and $t=b-1$ where $b$ is the current active learning round.

\begin{samepage}
\begin{algorithmblock}{FALL}\label{alg:FALL}
\algtext{\textbf{Input:} HSI $X=\{x_i\}_{i=1}^{N}$; labeled set $\mathcal{L}$; labels $\{y_i\}_{i\in\mathcal{L}}$; classes $\{1,\ldots,C\}$; Regularization $\{\tau_t\}_{t=1}^{B}$; exponent $p$; Budget $B$; Oracle $\mathcal{O}$; Initial Labels $\mathcal{L}_0$;}
\algtext{\textbf{Output:} Predicted labels $\widehat{Y}$}
\algline{Compute $\ell_p(x_i,x_j)$ for $i,j=1,\ldots,N$ and form weight matrix $W^{(p)}$.}
\algline{Initialize $\mathcal{L}_{\mathrm{curr}}\leftarrow\mathcal{L}_0$.}
\algline{\textbf{for} $b=1:B$ \textbf{do}}
\algline[1.5em]{Solve the graph Poisson system with source induced by $\mathcal{L}_{\mathrm{curr}}$ to obtain $\gamma$.}
\algline[1.5em]{Reweight the graph $(W_\gamma^{(p)})_{ij}=\gamma_i\gamma_jW_{ij}^{(p)}$.}
\algline[1.5em]{Run PWLL-$\tau$ using weights $W_\gamma^{(p_{\mathrm{true}})}$, labels $\mathcal{L}_{\mathrm{curr}}$ and $\tau_b$ weighting to obtain scores $u:X\rightarrow\mathbb{R}^{C}$.}
\algline[1.5em]{Select $i^\star=\arg\min_{i\notin\mathcal{L}}\|u(x_i)\|$.}
\algline[1.5em]{Query $y_{i^\star}=\mathcal{O}(x_{i^\star})$ from $\mathcal{L}$.}
\algline[1.5em]{Update $\mathcal{L}_{\mathrm{curr}}\leftarrow\mathcal{L}_{\mathrm{curr}}\cup\{x_{i^\star}\}$.}
\algline{Run PWLL-$\tau$ with final labels $\mathcal{L}_{\mathrm{curr}}$ and $\tau_B$ weighting for $u:X\rightarrow\mathbb{R}^{C}$ and assign $\hat{y}_i=\arg\max_k u_k(x_i)$.}
\algline{\textbf{return} $\widehat{Y}=\{\hat{y}_i\}_{i=1}^{N}$.}
\end{algorithmblock}

\begin{algorithmblock}{A-FALL}\label{alg:AFALL}
\algtext{\textbf{Input:} HSI $X=\{x_i\}_{i=1}^{N}$; labeled set $\mathcal{L}$; candidate set $\mathcal{P}$; landmarks $m$; MDS dimension $r$; budget $B$; update period $T$; oracle $\mathcal{O}$; initial exponent $p_0$; regularization $\{\tau_b\}_{b=1}^{B}$; Initial Labels $\mathcal{L}_0$;}
\algtext{\textbf{Output:} Predicted labels $\widehat{Y}$}
\algline{Select landmarks $Z=\{z_a\}_{a=1}^{m}\subset X$ using FPS.}
\algline{\textbf{for} $p\in\mathcal{P}$ \textbf{do}}
\algline[1.5em]{Form $D$ from distances $\ell_p(x_i,z_j)$ for each $i=1,\ldots,N$ and $j=1,\ldots,m$.}
\algline[1.5em]{Form self-tuned $k$NN weight matrix $W^{(p)}$ by computing embedding coordinates $\{x_i^{MDS}\}_{i=1}^{N}$.}
\algline{Initialize $p_{\mathrm{curr}}\leftarrow p_0$, $\mathcal{L}_{\mathrm{curr}}\leftarrow\mathcal{L}_0$.}
\algline{\textbf{for} $b=1:B$ \textbf{do}}
\algline[1.5em]{\textbf{if} $b\geq T$, $b\bmod T=0$, and $b<B$ \textbf{then}}
\algline[3em]{\textbf{for} $p\in\mathcal{P}$ \textbf{do}}
\algline[4.5em]{Form $S_p$ and $F_{p,a}^{(\mathcal{R}_a)}$ for each $a\in\mathcal{L}_{\mathrm{curr}}$.}
\algline[4.5em]{Compute $\mathbf{S}_{\mathrm{ALOO}}(p)$.}
\algline[3em]{Set $p_{\mathrm{curr}}\leftarrow\arg\max_{p\in\mathcal{P}}\mathbf{S}_{\mathrm{ALOO}}(p)$.}
\algline[1.5em]{Run PWLL-$\tau$ using weights $W^{(p_{\mathrm{curr}})}$, labels $\mathcal{L}_{\mathrm{curr}}$ and $\tau_b$ weighting to obtain scores $u:X\rightarrow\mathbb{R}^{C}$.}
\algline[1.5em]{Select $i^\star=\arg\min_{i\notin\mathcal{L}}\|u(x_i)\|$.}
\algline[1.5em]{Query $y_{i^\star}=\mathcal{O}(x_{i^\star})$ from $\mathcal{L}$.}
\algline[1.5em]{Update $\mathcal{L}_{\mathrm{curr}}\leftarrow\mathcal{L}_{\mathrm{curr}}\cup\{x_{i^\star}\}$.}
\algline{Run PWLL-$\tau$ with final labels $\mathcal{L}_{\mathrm{curr}}$ and $\tau_B$ weighting for $u:X\rightarrow\mathbb{R}^{C}$ and assign $\hat{y}_i=\arg\max_k u_k(x_i)$.}
\algline{\textbf{return} $\widehat{Y}=\{\hat{y}_i\}_{i=1}^{N}$.}
\end{algorithmblock}
\end{samepage}

\section{Experimental Results}\label{sec:results}

We show the improved experimental accuracy of FALL and A-FALL on the Salinas A HSI and a subset of the Pavia University HSI scene \cite{gic_hsi_scenes}. FALL, A-FALL and PWLL-$\tau$ were implemented in Python using NumPy and SciPy. The full Salinas scene was captured in Salinas Valley, California, and has a 3.7-meter spatial resolution per pixel with 224 spectral bands. The ground truth has 16 classes and a scene size of $512\times217$ pixels. The Salinas A HSI is a subset of the Salinas dataset, with a scene size of $83\times86$ pixels and 6 classes. The Pavia University HSI was captured over the University of Pavia, Italy, by the ROSIS sensor. The Pavia University HSI has a scene size of $610\times340$ pixels with 103 spectral bands, and a spatial resolution of 1.3 meters per pixel. The ground truth has 9 classes. In our experiments, we use a Pavia University subset of size $75\times180$ with 9 classes, corresponding to rows 253 to 327 and columns 30 to 209 of the original HSI. The performance of our algorithms is assessed using Overall Accuracy (OA, total fraction of correctly labeled pixels) and Average Accuracy (AA, mean per-class accuracies).

We test FALL and A-FALL against PWLL-$\tau$. We set $\tau_0=10^{-3}$, $\epsilon=10^{-9}$, and the $p$-learning update period $T=10$ in experiments. For A-FALL, we set $\mathcal{P}=\{1.5,2,3,4,6,8,10,12\}$ and $p_0=10$ for Salinas A. Similarly, we set $\mathcal{P}=\{1.25,1.5,1.75,2,2.25,2.5,3,4,6,8,10,12\}$ and $p_0=4$ for the Pavia subset. Optimal values for $p_0$ were found via grid search. We set $\mathcal{L}_0=\{x_R\}$ where $x_R$ is a randomly selected non-background point in $X$. Experiments were run on an ASUS ROG Strix G15 with 16GB of memory and an AMD Ryzen 7 4800H processor.

\begin{figure}[H]
\centering
\begin{subfigure}[t]{0.48\textwidth}
\includegraphics[width=\linewidth]{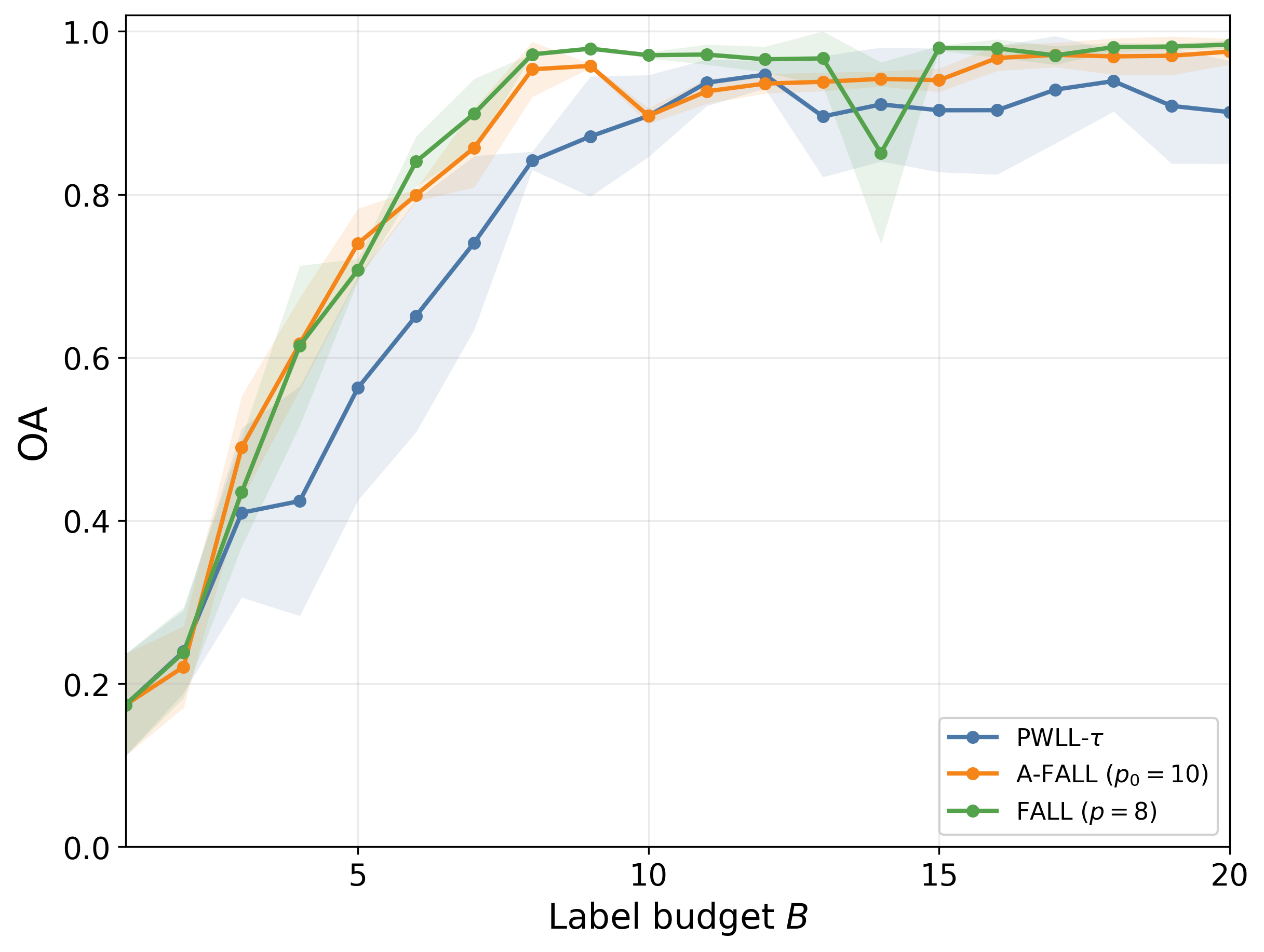}
\caption{OA}
\end{subfigure}
\hfill
\begin{subfigure}[t]{0.48\textwidth}
\includegraphics[width=\linewidth]{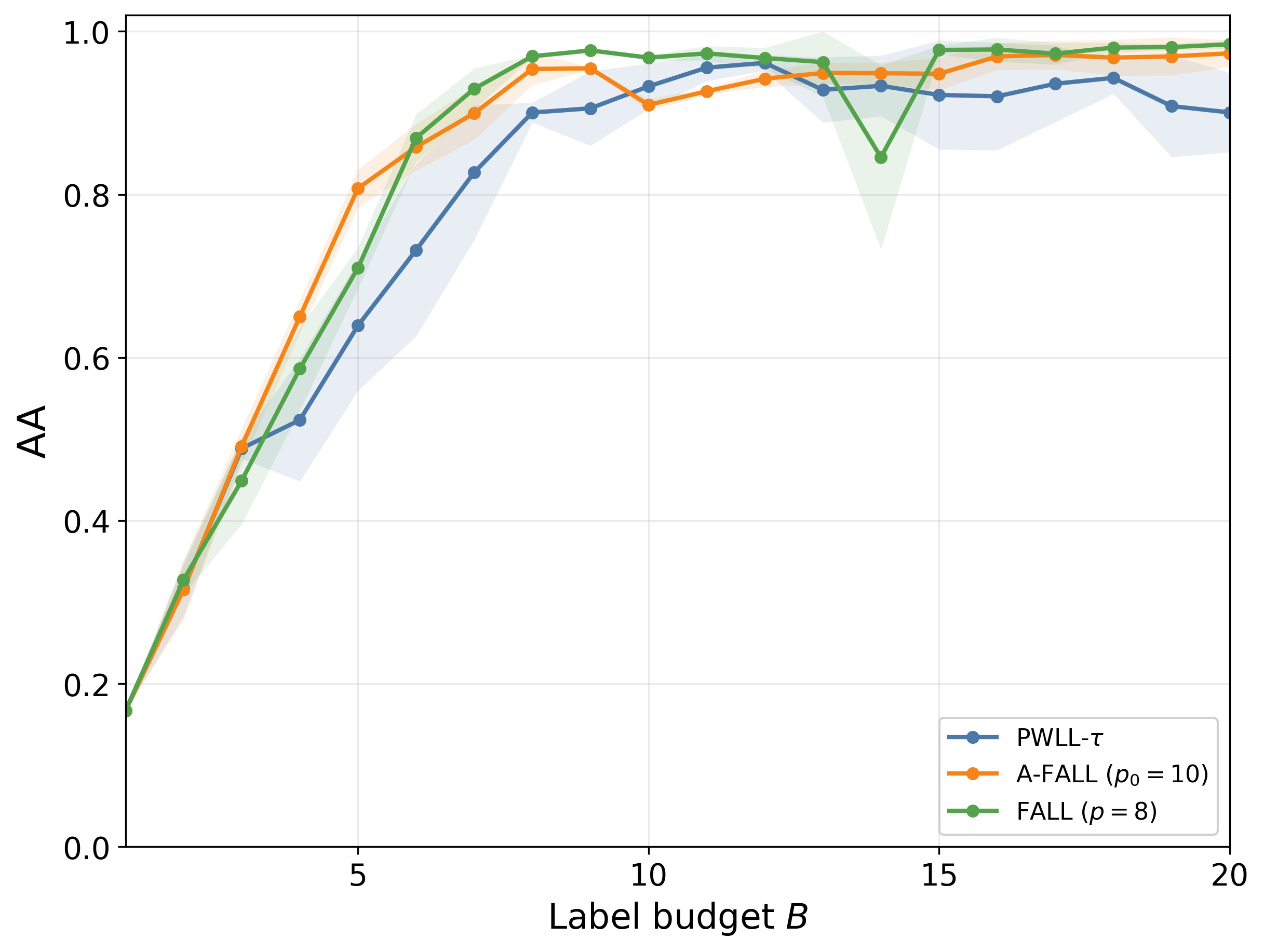}
\caption{AA}
\end{subfigure}
\caption{Mean OA and AA from $B=1$ to $B=20$ on Salinas A. Shaded regions show 1 standard deviation across seeds.}
\label{fig:salinas_curves}
\end{figure}

\begin{figure}[H]
\centering
\begin{subfigure}[t]{0.48\textwidth}
\includegraphics[width=\linewidth]{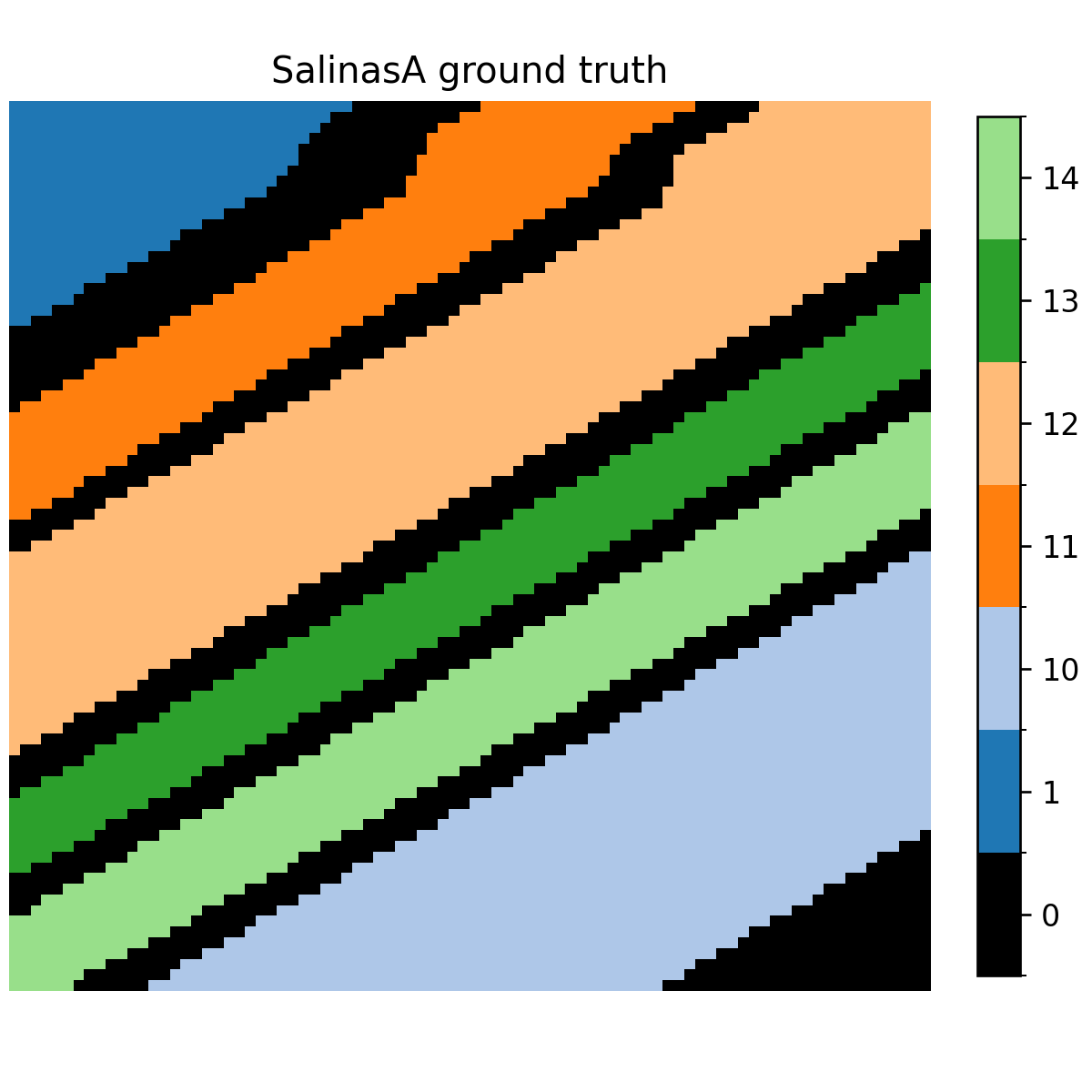}
\caption{Ground truth}
\end{subfigure}
\hfill
\begin{subfigure}[t]{0.48\textwidth}
\includegraphics[width=\linewidth]{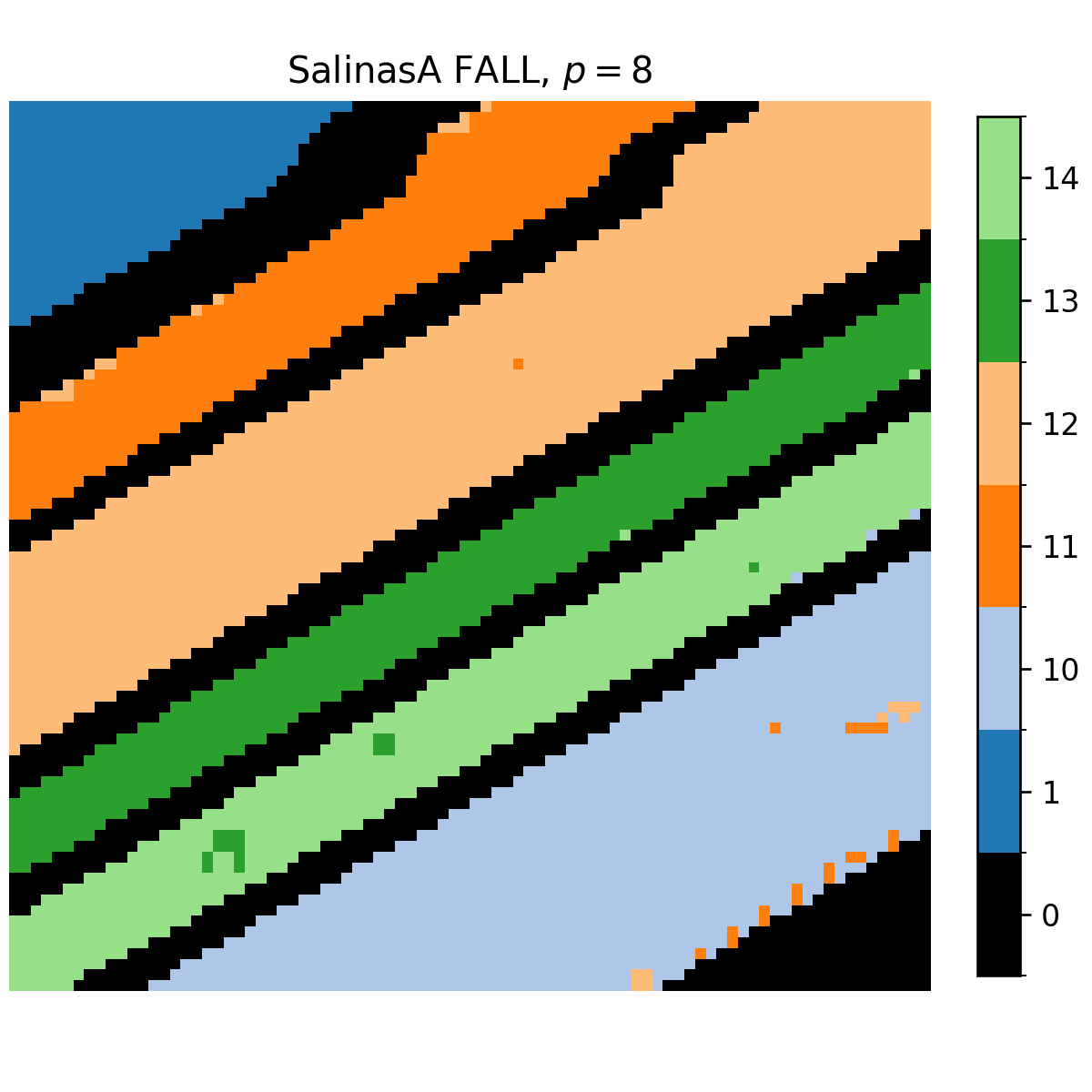}
\caption{FALL prediction}
\end{subfigure}
\caption{Salinas A GT vs FALL for $p=8$, $B=20$, best seed.}
\label{fig:salinas_maps}
\end{figure}

Table \ref{tab:results} and Figure \ref{fig:salinas_curves} show the efficacy of FALL and A-FALL over PWLL-$\tau$, with visualizations in Figure \ref{fig:salinas_maps}. FALL achieves the best OA and AA on Salinas A, while A-FALL obtains similar OA and AA scores and takes only 23.45 seconds to run on average. On Pavia, A-FALL obtains the highest OA and AA with the smallest average runtime of 93.48 seconds, demonstrating its scalability to larger HSI scenes.

\begin{table}[h]
\centering
\caption{Comparison of PWLL-$\tau$, FALL, and A-FALL on Salinas A at $B=20$ and the PaviaU crop at $B=30$, averaged over 10 seeds. Time is averaged over 10 seeds, measured in seconds. $\pm$ represents 1 standard deviation across seeds.}
\label{tab:results}
\begin{tabular}{lcccc}
\toprule
Method & $p/p_0$ & OA & AA & Time\\
\midrule
\multicolumn{5}{l}{\textit{Salinas A, $B=20$}}\\
\midrule
PWLL-$\tau$ & -- & $0.9010\pm0.0600$ & $0.9006\pm0.0462$ & \textbf{11.67}\\
FALL & $p=8$ & $\mathbf{0.9837\pm0.0053}$ & $\mathbf{0.9841\pm0.0047}$ & 37.40\\
A-FALL & $p_0=10$ & $0.9753\pm0.0159$ & $0.9731\pm0.0163$ & 23.45\\
\midrule
\multicolumn{5}{l}{\textit{PaviaU Subset, $B=30$}}\\
\midrule
PWLL-$\tau$ & -- & $0.8416\pm0.0366$ & $0.6252\pm0.0443$ & 130.54\\
FALL & $p=8$ & $0.8918\pm0.0358$ & $0.5987\pm0.0685$ & 197.21\\
A-FALL & $p_0=4$ & $\mathbf{0.9055\pm0.0300}$ & $\mathbf{0.6307\pm0.0586}$ & \textbf{93.48}\\
\bottomrule
\end{tabular}
\end{table}

\begin{figure}[H]
\centering
\begin{subfigure}[t]{0.48\textwidth}
\includegraphics[width=\linewidth]{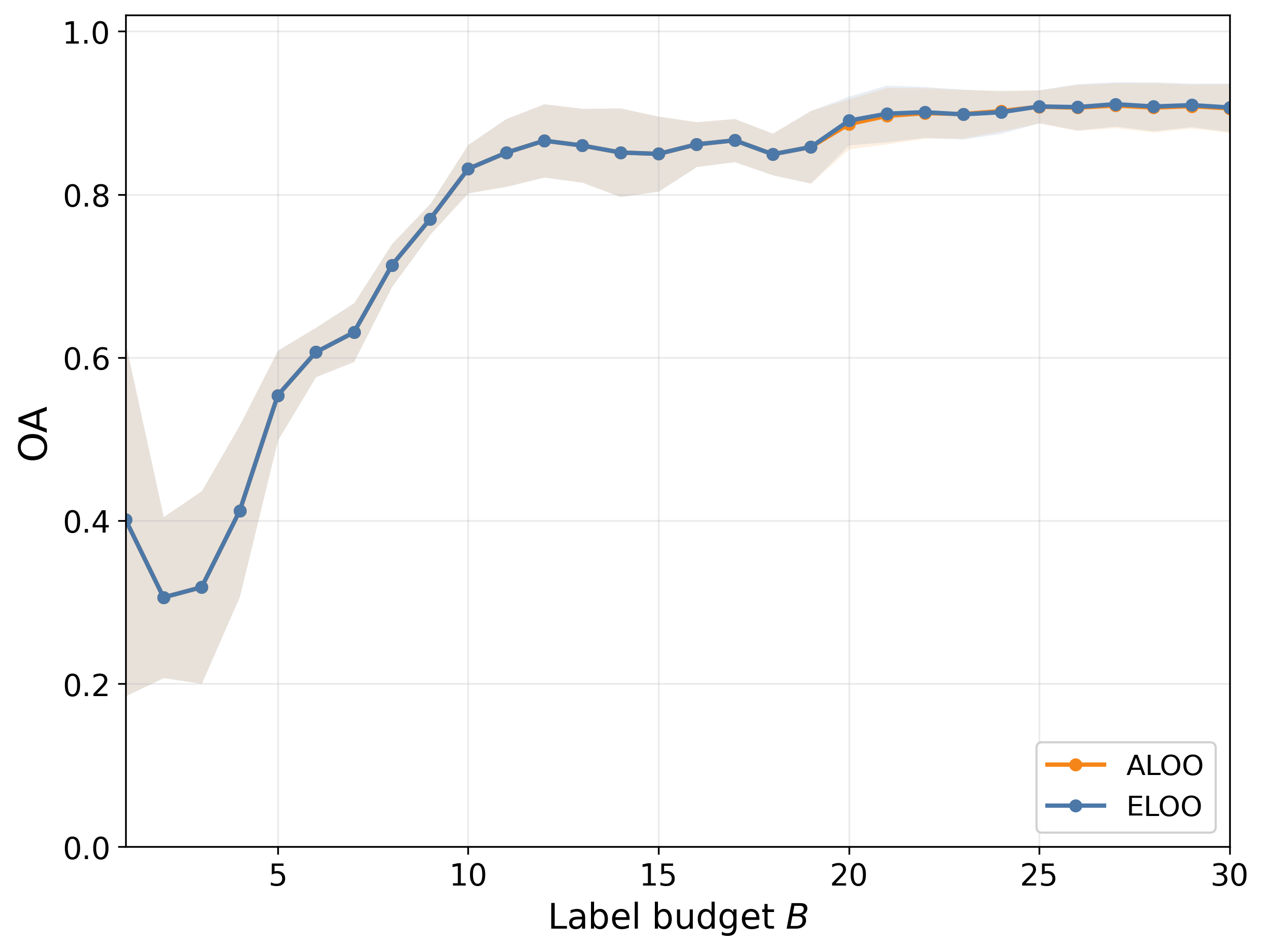}
\caption{OA}
\end{subfigure}
\hfill
\begin{subfigure}[t]{0.48\textwidth}
\includegraphics[width=\linewidth]{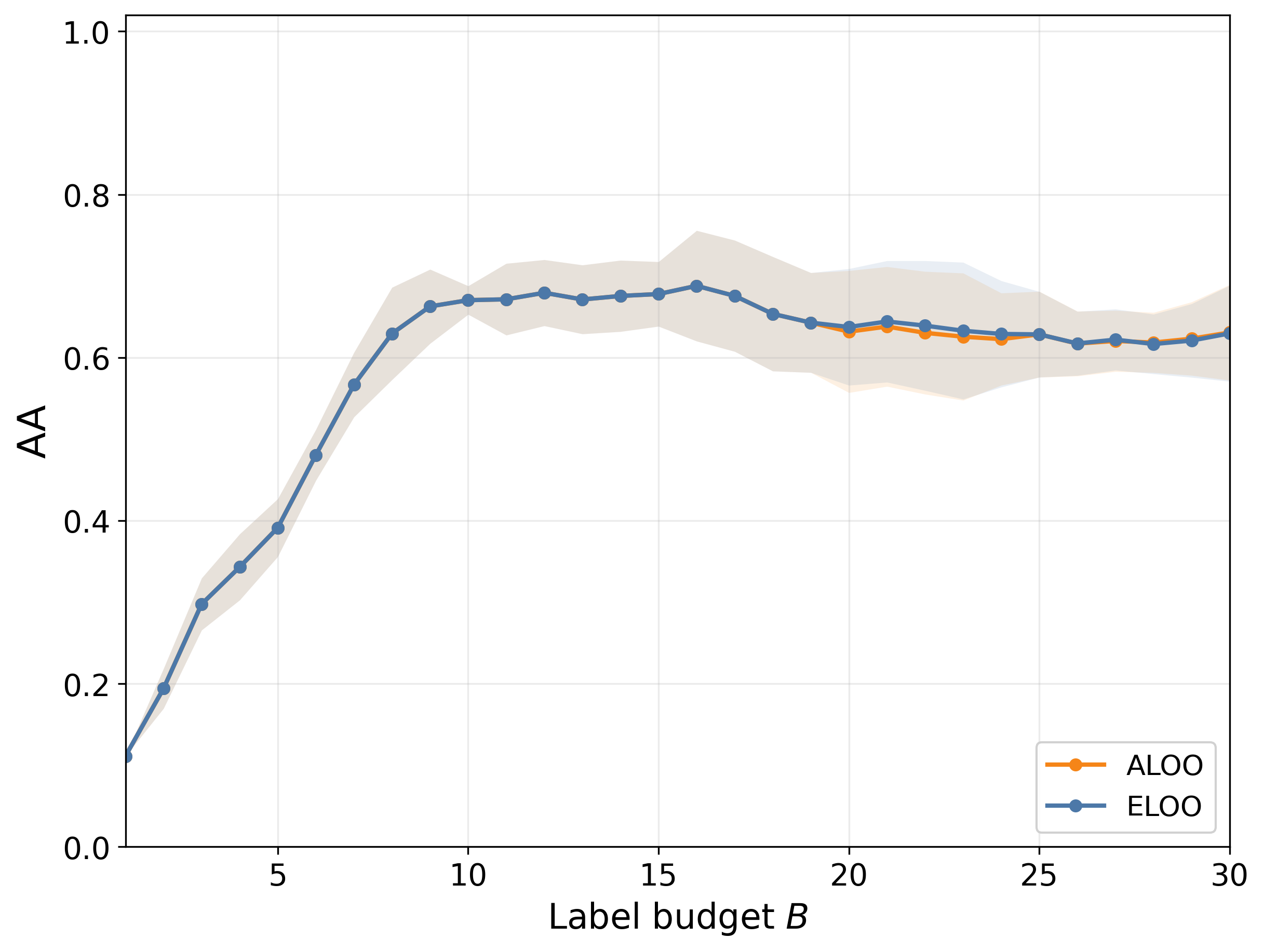}
\caption{AA}
\end{subfigure}
\caption{A-FALL with ELOO vs ALOO $p$-learning on the PaviaU subset from $B=1$ to $B=30$ over 10 seeds.}
\label{fig:pavia_eloo_aloo}
\end{figure}

We provide comparisons of ELOO and ALOO $p$-learning on the Pavia subset and display the differences in OA and AA in Figure \ref{fig:pavia_eloo_aloo}. At $B=30$, ELOO achieved an OA of $0.9066\pm0.0297$ and AA of $0.6296\pm0.0589$ with a mean runtime 252.59 s, while ALOO achieved an OA $0.9055\pm0.03$ and AA of $0.6307\pm0.0586$ with a mean runtime of 93.48 s.

\section{Conclusions and Future Work}

The proposed algorithms, FALL and A-FALL, provide improvements in overall accuracy compared to PWLL-$\tau$ in the low-label regime. Both algorithms provide a density-aware representation of the data manifold, with A-FALL using approximate Fermat distances via LMDS for greater efficiency. We introduce two methods to learn the Fermat exponent $p$ in our algorithm through cross-validation. In future work, we hope to explore further MDS variants that better preserve local Fermat distances and consider continuous or gradient-based $p$-learning methods. Further theoretical analysis on how using Fermat distances influences active class discovery is another direction of interest.

\bibliographystyle{unsrt}
\bibliography{refs}

\end{document}